\pdfoutput=1

\documentclass[11pt]{article}

\usepackage[preprint]{acl}

\usepackage{times}
\usepackage{latexsym}

\usepackage[T1]{fontenc}

\usepackage[utf8]{inputenc}

\usepackage{microtype}

\usepackage{inconsolata}

\usepackage{graphicx}

\usepackage{booktabs}
\usepackage{multirow}
\usepackage{enumitem}
\usepackage{longtable}
\usepackage{color}
\usepackage{colortbl}
\usepackage{amsmath}
\usepackage{makecell}

\title{Do RAG Systems Cover What Matters? Evaluating and Optimizing Responses with Sub-Question Coverage}

\author{Kaige Xie$^{1}$\thanks{Work done while interning at Salesforce AI Research.} ~~~Philippe Laban$^{2}$ ~~~Prafulla Kumar Choubey$^{2}$ \\ {\bfseries~~~Caiming Xiong$^{2}$ \quad ~~~Chien-Sheng Wu$^{2}$} \\
$^{1}$Georgia Institute of Technology ~~~ $^{2}$Salesforce AI Research \\
$^{1}$\texttt{kaigexie@gatech.edu} \\
$^{2}$\texttt{\{plaban, pchoubey, cxiong, wu.jason\}@salesforce.com} \\}

\begin{document}
\maketitle
\begin{abstract}
Evaluating retrieval-augmented generation (RAG) systems remains challenging, particularly for open-ended questions that lack definitive answers and require coverage of multiple sub-topics.
In this paper, we introduce a novel evaluation framework based on sub-question coverage, which measures how well a RAG system addresses different facets of a question.
We propose decomposing questions into sub-questions and classifying them into three types---core, background, and follow-up---to reflect their roles and importance.
Using this categorization, we introduce a fine-grained evaluation protocol that provides insights into the retrieval and generation characteristics of RAG systems, including three commercial generative answer engines: You.com, Perplexity AI, and Bing Chat.
Interestingly, we find that while all answer engines cover core sub-questions more often than background or follow-up ones, they still miss around 50\% of core sub-questions, revealing clear opportunities for improvement.
Further, sub-question coverage metrics prove effective for ranking responses, achieving 82\% accuracy compared to human preference annotations.
Lastly, we also demonstrate that leveraging core sub-questions enhances both retrieval and answer generation in a RAG system, resulting in a 74\% win rate over the baseline that lacks sub-questions.
\end{abstract}

\section{Introduction}

\begin{figure*}[t]
\centering
\includegraphics[width=\linewidth]{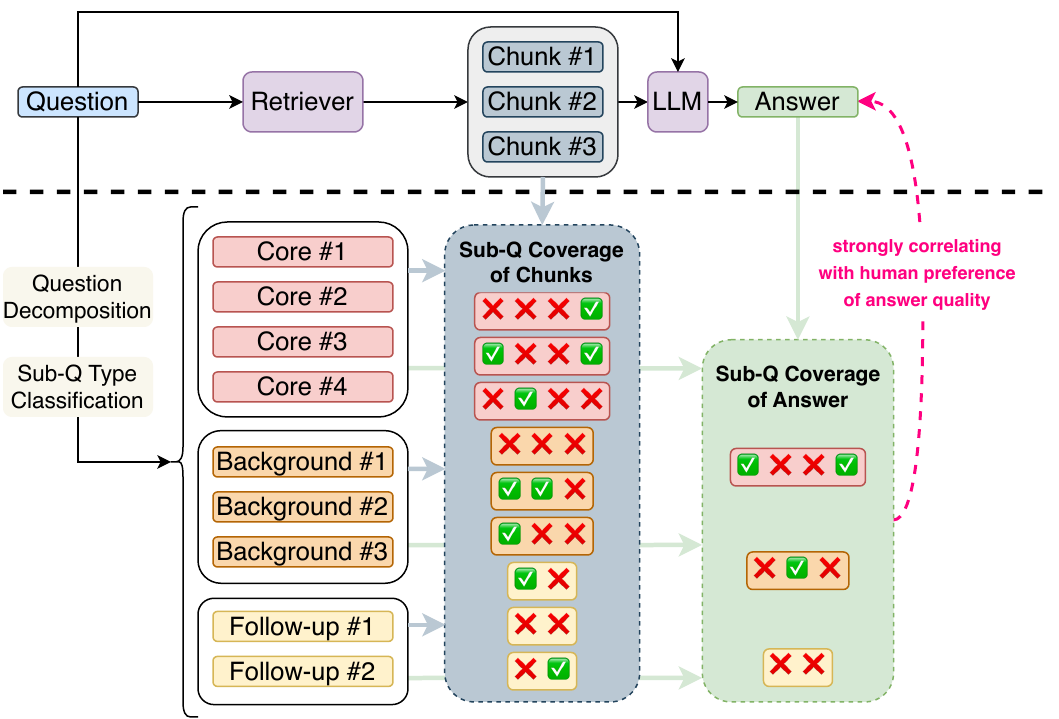}
\caption{An overview of our RAG evaluation framework based on \textit{sub-question coverage}.
Given a question, RAG's answer, and retrieved chunks, we decompose the question into sub-questions and classify them into three types: core, background, and follow-up;
we measure the sub-question coverage rates of both the answer and retrieved chunks with categorized sub-questions and design a fine-grained evaluation protocol to assess three popular RAG-based answer engines (\autoref{sec:evaluation}).
We find the sub-question coverage as an answer quality metric can approximate human perception of answer quality well (\autoref{sec:metric}).
We incorporate core sub-questions into different stages of the RAG workflow and effectively improve its responses (\autoref{sec:core-improve}).
}
\label{fig:intro}
\end{figure*}

Retrieval-augmented generation (RAG) has emerged as a powerful solution for answering open-ended complex questions by combining the strengths of retrieval systems with the generative capabilities of large language models (LLMs).
By retrieving relevant documents or knowledge chunks from external sources and using them as additional context when generating responses, RAG-based models can provide more accurate and contextually grounded answers compared to models that rely solely on generative approaches.
However, the evaluation of RAG systems remains a challenging task, particularly in the context of open-ended non-factoid questions where the strict correctness of a long-form answer is difficult to define~\cite{wang2024evaluating,han2024rag,rosenthal2024clapnq}.

One of the key issues in evaluating RAG systems is the lack of a systematic approach to assess how well these models address the full scope of complex questions.
Most existing RAG evaluations~\cite{yu2024evaluation,es-etal-2024-ragas,saad-falcon-etal-2024-ares} focus on surface-level metrics like faithfulness or relevance, without considering whether the generated response adequately covers the multi-dimensional nature of the question.
Complex questions are rarely monolithic in real-world applications.
They often require answers that address multiple facets of the issue at hand, which can be thought of as sub-questions that together form a comprehensive answer.
For instance, to fully answer the question ``How climate change is affecting the earth?'', we need to address multi-faceted information such as ``What are the impacts of rising temperatures on polar ice caps and glaciers?'' and ``What effects do the oceanic changes have on marine life?''.
However, existing approaches~\cite{LiuLlamaIndex2022,rosset2024researchy} leverage sub-questions as additional signals without fully considering how their relevance to the original query impacts response quality and user preferences.

This paper proposes a novel evaluation framework that introduces the concept of \textit{sub-question coverage} as a key metric for assessing RAG performance (depicted in \autoref{fig:intro}).
We hypothesize that complex, open-ended questions can be decomposed into several interrelated sub-questions, and they can be further classified into \textit{core} sub-questions that are central to addressing the main query, \textit{background} sub-questions that provide necessary context or supplementary information, and \textit{follow-up} sub-questions that may emerge from the original question but are not strictly required to formulate a satisfactory answer.
Some sub-questions, particularly follow-up or irrelevant background ones, may even be extraneous or detrimental to the quality of the response by diverting focus from the core query.
By utilizing this sub-question classification, we link the performance of RAG systems to the distribution of coverage across these three sub-question types, offering a more detailed assessment of the quality of the long-form answers.

There are currently no established datasets or frameworks for defining and classifying relevant sub-questions.
Although LLMs have been used for question decomposition, directly prompting them to generate sub-questions of each type results in low diversity and limited agreement with human classifications.
To address these challenges, we propose a two-step approach: 1. we decompose complex, open-ended, non-factoid questions into a comprehensive set of sub-questions; 2. we classify these sub-questions into core, background, and follow-up categories based on their functional roles within the main question.
This two-step prompting technique enhances the diversity of the generated sub-questions and achieves substantial agreement among human annotators and between human annotators and LLMs on type classification.

Next, we propose a fine-grained evaluation protocol based on sub-question coverage and analyze popular RAG-based answer engines---You.com\footnote{\url{https://you.com/}}, Perplexity AI\footnote{\url{https://www.perplexity.ai/}}, and Bing Chat\footnote{\url{https://www.bing.com/chat}}.
We evaluate two key aspects: (1) the patterns of retrieval and answer generation coverages across different sub-question types, and (2) the areas where search engines fall short, identifying opportunities for improvement.
We find that all three answer engines tend to prioritize core sub-questions over background or follow-up ones, which is the ideal behavior for generating high-quality answers.
However, our analysis also reveals key shortcomings.
None of the engines consistently retrieved knowledge chunks that fully addressed all core sub-questions.
Additionally, even when relevant core information was retrieved, the engines often failed to effectively utilize it in their generated answers.
This reveals their key limitations in both retrieval completeness and the use of retrieved content to produce comprehensive responses.

In addition to conducting a fine-grained analysis of answer comprehensiveness, we propose using sub-question classification to assess overall answer quality.
Leveraging human preference data from the WebGPT Comparisons dataset~\cite{nakano2021webgpt}, we find that addressing core sub-questions correlates most positively with human preferences, while background sub-questions show a moderate correlation.
In contrast, follow-up sub-questions negatively impact perceived answer quality.
Based on these findings, we introduce a weighted metric for evaluating answers that outperforms the conventional LLM-as-a-judge approach \cite{zheng2023judging} and strongly correlates with human preferences for open-ended answers.

Both our studies reveal how sub-question types can enhance the analysis of retrieval coverage and answer comprehensiveness.
A natural extension of this framework is improving RAG systems by aligning retrieval and answer generation with core sub-questions.
We evaluate four strategies for incorporating sub-question type information into different stages of RAG systems, demonstrating that the most effective approach uses core sub-questions during chunk retrieval and reranking.
Our results show that prioritizing core information significantly improves response quality, resulting in more accurate and comprehensive answers.

\section{Related Work}

\subsection{RAG Evaluation}
Existing work has proposed frameworks and benchmarks to evaluate RAG systems from different perspectives~\cite{gao2023retrieval,yu2024evaluation}.
ARES~\cite{saad-falcon-etal-2024-ares} focuses on context relevance, answer faithfulness, and relevance using lightweight language model judges.
RAGAS~\cite{es-etal-2024-ragas} offers a reference-free framework for evaluating retrieval and generation components.
RGB~\cite{chen2024benchmarking} benchmarks RAG on dimensions like noise robustness and information integration.
\citet{liu-etal-2023-evaluating} examine verifiability issues in generative search engines, highlighting concerns about citation accuracy.
\citet{cuconasu2024power} study retrieval strategies, focusing on passage relevance and positioning.
\citet{guinetautomated} propose automated task-specific accuracy measurement with synthetically generated exams.
RetrievalQA~\cite{zhang-etal-2024-retrievalqa} benchmarks adaptive RAG for short-form open-domain QA covering new world and long-tail knowledge.
SummHay~\cite{laban2024summary} evaluates long-context tasks with a focus on reference answer coverage and citation quality.
In contrast, our work evaluates RAG on complex, open-ended non-factoid questions through sub-question coverage and introduces a new automatic answer quality metric that strongly correlates with human preference.

\subsection{Query Optimization in RAG}

There are several ways to optimize input queries~\cite{gao2023retrieval} in RAG systems to enhance both retrieval and generation.
Key methods include query expansion and rewriting~\cite{jagerman2023query,amplayo-etal-2023-query,wang-etal-2023-query2doc} which align with prompting techniques like least-to-most prompting~\cite{zhou2023leasttomost}, decomposed prompting~\cite{khot2023decomposed}, and step-back prompting~\cite{zheng2024take}.
These techniques are particularly valuable for knowledge-intensive QA tasks requiring complex multi-hop reasoning~\cite{tang2024multihop,rosset2024researchy}.
RQ-RAG~\cite{chan2024rq} refines, decomposes, and disambiguates queries for multi-hop QA, while IRCoT~\cite{trivedi-etal-2023-interleaving} interleaves retrieval with chain-of-thought reasoning to reduce hallucinations.
\citet{ma-etal-2023-query} introduce a rewrite-retrieve-read framework, and Iter-RetGen~\cite{shao-etal-2023-enhancing} iteratively refines queries with generated outputs.
MULL~\cite{jia-etal-2024-mill} addresses ambiguous queries with a zero-shot query expansion framework, and Adaptive-RAG~\cite{jeong-etal-2024-adaptive} adapts retrieval strategies based on query complexity.
In contrast, our work focuses on decomposing complex open-ended non-factoid questions into sub-questions, leveraging them to evaluate and improve RAG systems.

\section{Sub-Question Coverage: A New Perspective for RAG Evaluation}
\label{sec:evaluation}

In this work, we focus on the RAG evaluation on long-form answers to open-ended and non-factoid questions~\cite{rosset2024researchy}, instead of questions with short and factoid answers.
Such questions are typically more complex, requiring deeper exploration, analysis, and synthesis of information.
They often lack a single correct answer, inviting subjective interpretations or multiple valid perspectives.
The integration of diverse sources are essential for forming a well-rounded response.

For instance, to address the question \textit{``Are fresh or frozen vegetables healthier?''} sufficiently, we might want to explore several sub-questions such as \#1 \textit{``How does the freezing process affect the nutritional content of vegetables?''}, \#2 \textit{``What are the common methods used to freeze vegetables?''}, and \#3 \textit{``What are the cost and taste differences between fresh and frozen vegetables?''}.
Essentially, the multi-faceted information necessary for answering the given question is equivalent to the overall information that can be covered by multiple sub-questions.
Nowadays, practical RAG systems~\cite{LiuLlamaIndex2022} usually involve this general query decomposition step to improve their response quality and coverage.
However, while gathering more information from these various sub-questions can be beneficial, we argue that not all should be treated equally, as their relevance and importance to the original question may vary.
For example, sub-question \#1 is the most crucial, \#2 provides helpful context, and \#3 encourages thinking one step ahead.

Motivated by this observation, we propose a new approach for automatic RAG evaluation: \textit{sub-question coverage}.
We first decompose an open-ended non-factoid question into sub-questions, classifying them as core, background, or follow-up (\S \ref{subsec:decomposition}), based on their relevance and functional role in answering the main question.
Using this taxonomy, we develop a fine-grained evaluation protocol with metrics to assess the multi-dimensional effectiveness of RAG systems (\S \ref{subsec:protocol}).
Since retrieval is key in RAG, our protocol examines the coverage of sub-questions in both the long-form answers and retrieved chunks.
This allows us to analyze how retrieval affects sub-question coverage in the final answer.
We then apply this evaluation to three popular RAG systems: You.com, Perplexity AI, and Bing Chat (\S \ref{subsec:result-1}).

In our experiments, we manually select 200 open-ended non-factoid questions with a high level of complexity from the Researchy Questions dataset~\cite{rosset2024researchy} and perform a question decomposition on them to get all corresponding sub-questions and their types.
Our criteria for question selection is that the question must be multi-faceted and must require multi-hop reasoning to answer.
In total, we have 200 questions and around 4000 sub-questions.

\subsection{Question Decomposition}
\label{subsec:decomposition}

We prompt GPT-4~\cite{openai2023gpt4} to first come up with a comprehensive collection of relevant sub-questions that can answer the main question fully, and then prompt it again to classify sub-questions into three types: \textit{core}, \textit{background}, and \textit{follow-up} sub-questions.
These three sub-question types are defined as follows, and our prompts are provided in \autoref{tab:prompt-1}.
\begin{itemize}[leftmargin=*]
\setlength\itemsep{0.001em}
\item{\underline {\textit{Core Sub-Question:}}} A core sub-question is central to the main topic and directly or partially addresses the main question. It is crucial for interpreting the logical reasoning of the main question and provides essential insights required for answering it. These sub-questions often involve multiple steps or perspectives, making them fundamental to generating comprehensive and well-rounded responses.

\item{\underline {\textit{Background Sub-Question:}}}
A background sub-question is optional when answering the main question, but it can provide additional context or background information that helps clarify the main query. Its primary role is to support the understanding of the main topic by offering supplementary evidence or information, though it is not strictly necessary for addressing the core aspects of the question.

\item{\underline {\textit{Follow-up Sub-Question:}}}
A follow-up sub-question is not needed to answer the main question. These sub-questions often arise after users receive an initial answer and seek further clarification or details. They may explore specific aspects of the response in greater depth, but their answers can sometimes be out-of-scope or beyond the focus of the original query.
\end{itemize}

\paragraph{Human Annotation and Classification Accuracy of Sub-Questions:} We recruit five human annotators to manually classify 200 sub-questions into three types of sub-questions, and we take the majority vote of the annotations as the ground-truth type.
We first observe that human annotators have an average 74.6\% accuracy, implying that our task might involve some subjectivity depending on user expectations, as some sub-questions might be less important for some people but could be interesting knowledge for other users.
For example, we notice from the annotations that some people find the sub-question \textit{``What are the current levels of natural gas reserves?''} important for answering the main question \textit{``Why has the natural gas price increased recently?''} and classify it as a core sub-question, while others find it helpful but not as important and therefore classify it as a background sub-question.
We find that annotators have higher agreement on core and follow-up sub-questions with 76.8\% and 79.2\% accuracy respectively, while on background sub-questions their classification accuracy is only 64.3\%.
Surprisingly, we find that GPT-4's zero-shot classification accuracy is 77.5\%, and few-shot classification accuracy is 84.8\%.
This is a good indicator that we can rely on LLMs to automatically classify the sub-questions.
We provide samples of question decomposition with three-way classification in \autoref{tab:samples}.

\subsection{Fine-grained Evaluation Protocol}
\label{subsec:protocol}

The comprehensive question decomposition of different types can enable a fine-grained evaluation of RAG systems based on the sub-question coverage of both the long-form answers and the retrieved chunks.
Through this fine-grained evaluation, we seek to answer the following two central questions:
\begin{itemize}[leftmargin=*]
\setlength\itemsep{0.001em}
    \item What percentage of core, background and follow-up sub-questions are covered in the long-form answer?
    \item If a core, background or follow-up sub-question is not covered, is it because the RAG system fails to retrieve chunks that contain the necessary knowledge for covering the sub-question, or because the LLM used for response generation fails to identify and include the necessary knowledge from the retrieved chunks?
\end{itemize}

We prompt GPT-4 with few-shot annotated examples to automatically measure the sub-question coverage (prompt provided in \autoref{tab:prompt-2}).
Given a piece of text (either a long-form answer or a retrieved chunk) and a sub-question, we ask GPT-4 to judge if there exists any part of the given text that can answer the sub-question.
If GPT-4 believes the sub-question can be answered (i.e., covered), we ask it to further identify the text fragment that answers the sub-question.
To evaluate the reliability of this approach, we compared GPT-4's automatic judgments with human annotations on 100 samples, finding an 83\% alignment rate.
This strong agreement provides confidence in the accuracy of our automatic sub-question coverage measurement.
 
For each of the three sub-question types (denoted as $\textit{type} \in \{\text{core}, \text{background}, \text{follow-up}\}$), we calculate the percentage occurrence of each of the following four scenarios:

\begin{itemize}[leftmargin=*]
\setlength\itemsep{0.001em}
    \item $P_{\text{\textit{type}}}(\neg \text{answered}, \neg \text{retrieved})$: the sub-question is neither covered by the long-form answer nor by any of the retrieved chunks;
    \item $P_{\text{\textit{type}}}(\neg \text{answered}, \text{retrieved})$: the sub-question is not covered by the long-form answer, but is covered by at least one of the retrieved chunks;
    \item $P_{\text{\textit{type}}}(\text{answered}, \neg \text{retrieved})$: the sub-question is covered by the long-form answer, but is not covered by any of the retrieved chunks;
    \item $P_{\text{\textit{type}}}(\text{answered}, \text{retrieved})$: the sub-question is covered by both the long-form answer and at least one of the retrieved chunks.
\end{itemize}

We design four metrics based on the percentage occurrence of the above scenarios:
\begin{itemize}[leftmargin=*]
\setlength\itemsep{0.001em}
    \item \textbf{Metric \#1}: answer's sub-question coverage rate, expressed as $P_{\text{\textit{type}}}(\text{answered})$.
    \item \textbf{Metric \#2}: retrieval's sub-question coverage rate, expressed as $P_{\text{\textit{type}}}(\text{retrieved})$.
    \item \textbf{Metric \#3}: the capability to identify core knowledge from retrieved chunks, expressed as $\dfrac{P_{\text{core}}(\text{answered, retrieved})}{P_{\text{core}}(\text{retrieved})}$.
    \item \textbf{Metric \#4}: the potential of getting performance gain by improving retrieval for core sub-questions, expressed as $\dfrac{P_{\text{core}}(\neg \text{answered}, \neg \text{retrieved})}{P_{\text{core}}(\neg \text{answered})}$.
\end{itemize}

In addition, as RAG systems usually retrieve ten or more chunks as additional context for the LLM, we want to know when a core sub-question is covered / not covered in the long-form answer, on average what percentage of retrieved chunks ($p_{\text{covered}}$ and $p_{\text{not covered}}$) cover this core sub-question.
We design \textbf{Metric \#5}: the correlation between core sub-question coverage in the long-form answer and core knowledge's appearance frequency in all retrieved chunks.
This correlation is measured by the difference between these two percentage numbers: $p_{\text{covered}}-p_{\text{not covered}}$.
This correlation also indicates how good a RAG system is at prioritizing core knowledge in its long-form answer.

The identified text fragments in the automatic judgment of sub-question coverage allow us to locate the place in the long-form answer where it starts to address the given sub-question, which can be expressed by a percentage number: e.g. 20\% represents the 20th word of a 100-word answer.
We name this place ``addressing position'' ($\text{pos}_{\text{\textit{type}}}$), and use it to design \textbf{Metric \#6}: long-form answer's position alignment with human's writing habit---people generally prefer the core and background information to be at the beginning of an answer, with the follow-up information at the end.
This alignment is measured by the difference between the average addressing positions of follow-up and core/background sub-questions: $\text{pos}_{\text{follow-up}}-(\text{pos}_{\text{core}}+\text{pos}_{\text{background}})/2$.

\subsection{Evaluation of RAG-based Answer Engines}
\label{subsec:result-1}

\begin{table*}
\centering
\small
\begin{tabular}{r|ccc|ccc|ccc}
    \toprule
    \multicolumn{1}{r|}{\textbf{Answer Engine}} & \multicolumn{3}{c|}{\textsc{You.com}} & \multicolumn{3}{c|}{\textsc{Perplexity AI}} & \multicolumn{3}{c}{\textsc{Bing Chat}} \\
    \cmidrule(lr){1-10}
    \textbf{Sub-Question Type} & \textbf{C} & \textbf{B} & \textbf{F} & \textbf{C} & \textbf{B} & \textbf{F} & \textbf{C} & \textbf{B} & \textbf{F} \\
    \cmidrule(lr){1-10}
    $\neg \text{\textbf{answered}}, \neg \text{\textbf{retrieved}}$ & $26\%$ & $32\%$ & $56\%$ & $28\%$ & $39\%$ & $61\%$ & $26\%$ & $39\%$ & $59\%$ \\
    $\neg \text{\textbf{answered}}, \text{\textbf{retrieved}}$ & $32\%$ & $48\%$ & $30\%$ & $18\%$ & $41\%$ & $22\%$ & $25\%$ & $47\%$ & $32\%$ \\
    $\text{\textbf{answered}}, \neg \text{\textbf{retrieved}}$ & $9\%$ & $3\%$ & $4\%$ & $9\%$ & $3\%$ & $5\%$ & $7\%$ & $1\%$ & $2\%$ \\
    $\text{\textbf{answered}}, \text{\textbf{retrieved}}$ & $33\%$ & $17\%$ & $10\%$ & $45\%$ & $17\%$ & $12\%$ & $42\%$ & $13\%$ & $7\%$ \\
    \bottomrule
\end{tabular}
\caption{Three answer engines' percentage occurrences of four scenarios (based on sub-question coverage of both the long-form answer and the retrieved chunks) for three sub-question types core (C), background (B), and follow-up (F).}
\label{tab:percentages}
\end{table*}

\begin{table*}
\centering
\resizebox{\textwidth}{!}{
\begin{tabular}{l|ccc|c}
    \toprule
     & \textsc{You.com} & \textsc{Perplexity AI} & \textsc{Bing Chat} & \textbf{Ranking} \\
    \cmidrule(lr){1-5}
    \textbf{Metric \#1 (core)} & $42\%$ & $54\%$ & $49\%$ & \textsc{Perplexity AI} > \textsc{Bing Chat} > \textsc{You.com} \\
    \cmidrule(lr){1-5}
    \textbf{Metric \#1 (background)} & $20\%$ & $20\%$ & $14\%$ & \textsc{You.com} = \textsc{Perplexity AI} > \textsc{Bing Chat} \\
    \cmidrule(lr){1-5}
    \textbf{Metric \#1 (follow-up)} & $14\%$ & $17\%$ & $9\%$ & \textsc{Perplexity AI} > \textsc{You.com} > \textsc{Bing Chat} \\
    \cmidrule(lr){1-5}
    \textbf{Metric \#2 (core)} & $65\%$ & $63\%$ & $67\%$ & \textsc{Bing Chat} > \textsc{You.com} > \textsc{Perplexity AI} \\
    \cmidrule(lr){1-5}
    \textbf{Metric \#2 (background)} & $65\%$ & $58\%$ & $60\%$ & \textsc{You.com} > \textsc{Bing Chat} > \textsc{Perplexity AI} \\
    \cmidrule(lr){1-5}
    \textbf{Metric \#2 (follow-up)} & $40\%$ & $34\%$ & $39\%$ & \textsc{You.com} > \textsc{Bing Chat} > \textsc{Perplexity AI} \\
    \cmidrule(lr){1-5}
    \makecell[l]{\textbf{Metric \#3}} & $51\%$ & $71\%$ & $63\%$ & \textsc{Perplexity AI} > \textsc{Bing Chat} > \textsc{You.com} \\
    \cmidrule(lr){1-5}
    \makecell[l]{\textbf{Metric \#4}} & $45\%$ & $61\%$ & $51\%$ & \textsc{Perplexity AI} > \textsc{Bing Chat} > \textsc{You.com} \\
    \cmidrule(lr){1-5}
    \makecell[l]{\textbf{Metric \#5}} & $11\%$ & $53\%$ & $39\%$ & \textsc{Perplexity AI} > \textsc{Bing Chat} > \textsc{You.com} \\
    \cmidrule(lr){1-5}
    \makecell[l]{\textbf{Metric \#6}} & $36\%$ & $45\%$ & $60\%$ & \textsc{Bing Chat} > \textsc{Perplexity AI} > \textsc{You.com} \\
    \bottomrule
\end{tabular}
}
\caption{A fine-grained evaluation of three answer engines with Metrics \#1-6, as well as a ranking for each metric.}
\label{tab:metrics}
\end{table*}

We use the fine-grained evaluation protocol to assess three popular RAG-based answer engines: You.com, Perplexity AI, and Bing Chat.
When querying these answer engines, we ask them to give responses that have around 300 words.
On average, their generated responses have 272 words.
To acquire their retrieved documents, we collect the citation information and scrape the text of web pages that answer engines have used as knowledge sources for generating long-form answers.

The percentage occurrences of four scenarios for three sub-question types are presented in \autoref{tab:percentages}.
We also assess three answer engines with Metrics \#1-6.
The results are presented in \autoref{tab:metrics}.
In this way, we enable a fine-grained measurement of answer engines' multi-dimensional effectiveness and give insights into their pros and cons from the perspective of sub-question coverage.

First of all, these commercial answer engines indeed follow the trend that the answers to core sub-questions appear more often than the other two types.
For example, in \textbf{Metric \#1}, You.com has $9\%+33\%=42\%$ for core, $3\%+17\%=20\%$ for background, and $4\%+10\%=14\%$ for follow-up sub-questions.
A similar trend is observed in \textbf{Metric \#2} in terms of retrieving the knowledge for core sub-questions.
Also, when retrieved chunks have an answer (retrieved=yes), answers to core sub-questions have a higher chance of being shown in final responses.
For instance, in \textbf{Metric \#3}, You.com has about $33\%/(33\%+32\%)=51\%$ of the time included core answers; however, its background or follow-up sub-questions are covered only about $17\%/(17\%+48\%)\approx 10\%/(10\%+30\%)=25\%$ of the time.
The exhibited behavioral discrepancy across different sub-question types confirms that our hypothesis is appropriate and effective in distinguishing diverse relationships between sub-questions and the main question.

Based on \textbf{Metric \#4}, we observe that all answer engines have substantial potential for performance gains by improving retrieval for core sub-questions.
When comparing performance on \textbf{Metric \#5}, all three systems face some challenges in translating retrieved core knowledge into the final answer.
Perplexity AI is more successful in connecting its retrieval and generation phases, while You.com, with only 11\%, shows a significant gap in utilizing retrieved content.
To improve systems like You.com, enforcing the inclusion of core sub-questions during answer generation could significantly enhance response quality.
Finally, \textbf{Metric \#6} indicates that Bing Chat organizes the different types of information (core, background, and follow-up) in a more aligned and coherent way compared to the other two engines.
To further align these systems with the human writing style, enforcing a structured order of sub-question types in the response generation stage would help improve the overall flow and comprehensiveness of the answers.

\section{An Automatic Answer Quality Metric}
\label{sec:metric}

\begin{table}
\centering
\small
\begin{tabular}{l|c}
    \toprule
    \textbf{Metric} & \textbf{Accuracy} \\
    \midrule
    \textsc{LLM-as-a-judge} & 0.71 \\
    \textsc{Core Only} & 0.78 \\
    \textsc{All-Type Hybrid} & \textbf{0.82} \\
    \bottomrule
\end{tabular}
\caption{Three automatic answer quality metrics' prediction accuracy on 500 samples, each of which has a question, two responses, and a human preference for one of the two. A higher accuracy number indicates a stronger correlation with human preference.}
\label{tab:correlations}
\end{table}

The sub-question coverage enables a systematic evaluation of a RAG system from a developer's standpoint, which spans both retrieval and generation.
In contrast, end-users would perceive the effectiveness of a RAG system directly from the quality of answers the RAG system generates.
They may evaluate the answer quality based on a variety of criteria such as completeness and relevance.
While existing work~\cite{zheng2023judging} tries to automate this evaluation by approximating human preferences with the LLM as a judge, we believe that directly comparing two long answers poses a complex challenge for LLMs.
Therefore, by systematically identifying the types of sub-questions addressed in a given answer, we propose to streamline the answer evaluation process into a more robust framework.
To validate our approach, we investigate a reliable automatic answer quality metric derived from sub-question coverage and analyze its relationship with human preferences.

For our study, we select 500 non-factoid open-ended questions from the WebGPT Comparisons dataset, specifically targeting ``why'' and ``how'' questions that have long-form answers\footnote{\url{https://huggingface.co/datasets/openai/webgpt_comparisons}}~\cite{nakano2021webgpt}.
Each selected sample contains a question, two answers, and a preference score from humans (ranging from -1 to 1) indicating which of the two responses is preferred by humans.
We remove samples whose preference scores are equal to zero (meaning a tie), and map scores to preference labels (A > B or B > A) according to the sign of scores.
For each question, we perform our question decomposition and get a collection of core, background, and follow-up sub-questions.
For each sub-question, we use the automatic sub-question coverage judgment (introduced in \autoref{subsec:protocol}) to check whether a given response contains a corresponding answer.
This results in three coverage numbers (\%) based on three sub-question types for each response: $\{\text{c}_{\text{core}},\text{c}_{\text{background}},\text{c}_{\text{follow-up}}\}$.

We start by studying how well the core sub-question coverage rate, $\text{c}_{\text{core}}$, correlates with human preference.
We assume that if a response has a higher $\text{c}_{\text{core}}$ then it will be more preferable than the other response.
Results presented in \autoref{tab:correlations} show that our metric based on core sub-question coverage (denoted as \textsc{Core Only}) achieves 78\% accuracy, outperforming the random change (50\%) by a big margin.
It also surpasses \textsc{LLM-as-a-judge}~\cite{zheng2023judging}, which directly prompts GPT-4 with an instruction to make a pairwise comparison between two responses, suggesting the effectiveness of automatically evaluating answer quality from the perspective of core sub-question coverage.

Given the success in building an automatic metric from core sub-question coverage, we further study how the sub-question coverage of the other two types, $\text{c}_{\text{background}}$ and $\text{c}_{\text{follow-up}}$, may impact a metric's overall correlation with human preference.
To take three types of sub-question coverage into account simultaneously, we derive the response rating using a weighted sum of three coverage numbers, expressed as
$$
    \text{rating}=\sum_{\text{type}}\text{w}_{\text{type}}*\text{c}_{\text{type}},
$$
where $\text{w}_{\text{type}}$ represents the weighting coefficient.
We maintain a 100-sample hold-out validation set and perform a grid search on it to look for the best weighting coefficients according to the validation-set prediction accuracy.
The best weighting coefficients we found are  $\text{w}_{\text{core}}:\text{w}_{\text{background}}:\text{w}_{\text{follow-up}}=1:0.5:-1$, and the corresponding prediction accuracy is 82\% (denoted as \textsc{All-Type Hybrid}).
The ratios between weighting coefficients indicate that human preference correlates positively with the core sub-question, followed by the background sub-question, while negatively with the follow-up sub-question, demonstrating a stronger correlation with human preference and thus the great effectiveness of our automatic answer quality metric.
It’s important to note that human preferences can be subjective, varying across different user personas and use cases, which may affect these ratios.
Our work takes an initial step in exploring sub-question coverage as a potential reward signal, leaving further generalization experiments for future research.

\section{Improving RAG Responses with Core Sub-Question}
\label{sec:core-improve}

\begin{table}
\centering
\resizebox{\linewidth}{!}{
\begin{tabular}{c|ccccc}
    \toprule
    \textbf{Method} & \textsc{B} & \textsc{M1} & \textsc{M2} & \textsc{M3} & \textsc{M4} \\
    \midrule
    \textsc{B} & - & 41.5\% & 34\% & 26.75\% & 34.75\% \\
    \textsc{M1} & 58.5\% & - & 30.5\% & 25\% & 36.5\% \\
    \textsc{M2} & 66\% & 69.5\% & - & 35.75\% & 40.75\% \\
    \textsc{M3} & 73.25\% & 75\% & 64.25\% & - & 57.5\% \\
    \textsc{M4} & 65.25\% & 63.5\% & 59.25\% & 42.5\% & - \\
    \bottomrule
\end{tabular}
}
\caption{Win Rates between five methods: \textsc{Baseline} (\textsc{B}), \textsc{[Query-Augmentation]}$_{\text{definition}}$ (\textsc{M1}), \textsc{[Query-Augmentation]}$_{\text{subquestion}}$ (\textsc{M2}), \textsc{[Retrieval-Augmentation]} (\textsc{M3}), and \textsc{[E2E-Augmentation]} (\textsc{M4}). Each win rate indicates the percentage of times that the method in the row outperforms the method in the column.}
\label{tab:improvements}
\end{table}

The proven strong correlation between core sub-question coverage and human perception of answer quality motivates us to explore the feasibility of enhancing RAG responses by augmenting RAG with core sub-questions.
This augmentation essentially involves incorporating core sub-questions into the RAG system, which can happen in different stages of the RAG workflow, including query transformations, retrieval, and generation.
Specifically, we design the following techniques to incorporate core sub-questions:

\begin{itemize}[leftmargin=*]
\setlength\itemsep{0.001em}
    \item \textsc{[Query-Augmentation]}$_{\text{definition}}$ We augment the input query with the definition of core sub-questions (introduced in \autoref{subsec:decomposition}).
    We ask the RAG system to come up with core sub-questions of the main question and try to cover as many core sub-questions as possible when generating the response.
    This is an indirect approach since the same definition is applied uniformly to all questions, which does not necessarily enhance retrieval.
    However, it can be beneficial during response generation, as the language model is prompted to concentrate on the core sub-questions.
    \item \textsc{[Query-Augmentation]}$_{\text{subquestion}}$ We augment the input query directly with core sub-questions we get from the question decomposition (introduced in \autoref{subsec:decomposition}).
    We prompt the RAG system to address as many core sub-questions as possible in the generated response.
    This approach improves retrieval recall as core sub-questions are explicitly used during the retrieval process, and it also aids in generation, as the language model is directly instructed to provide answers for all core sub-questions.
    \item \textsc{[Retrieval-Augmentation]} We augment the retrieval process by separately retrieving the top relevant chunks for both the original query and all core sub-questions.
    These retrieved chunks are then combined into a unified pool, and reranked based on how well they cover the core sub-questions.
    After reranking, we select the top-K chunks and use them to generate a response to the original query.
    This approach increases the coverage of core sub-questions in the retrieved chunks.
    \item \textsc{[E2E-Augmentation]} We enhance both the retrieval and generation processes using core sub-questions.
    For each core sub-question, we first retrieve top-K relevant chunks and generate a corresponding answer.
    And then, we combine all the core sub-answers and then prompt the LLM to produce a final answer to the original query.
    This ensures that the final answer is explicitly informed by the detailed answers to each core sub-question.
\end{itemize}

\paragraph{Experimental Setup:} We implement our RAG system using LlamaIndex \cite{LiuLlamaIndex2022} and build the retrieval pool by concatenating all the cited sources gathered from the answer engines discussed in \autoref{subsec:result-1}.
We use the same set of 200 open-ended, non-factoid questions and use the VectorStoreIndex\footnote{\url{https://docs.llamaindex.ai/en/stable/module_guides/indexing/vector_store_index/}} with the OpenAI's \texttt{text-embedding-ada-002} model for embeddings.
We set the top-K to 10 for retrieval and generate a long-form answer of around 300 words.

We evaluate the baseline and core sub-question-informed RAG systems using a win-rate matrix.
Each pair of responses is compared using the GPT-4 Judge, and we report the percentage of times one method outperforms the other.
To remove position bias in the pairwise comparison, each pair is evaluated twice by swapping their order.
We rely on GPT-4 Judge as the evaluator, despite it being less effective than our proposed answer quality metrics, because it is widely used in prior works and evaluation libraries.
Additionally, since both our RAG enhancement approaches and answer quality metrics are based on core sub-questions, using our own framework for evaluation would have introduced bias.
The results are presented in \autoref{tab:improvements}.

Our results show that all systems incorporating core sub-questions outperform the baseline, demonstrating that incorporating core sub-questions at any stage in the RAG pipeline is effective.
Notably, Retrieval Augmentation achieves the highest win rates across all comparisons, outperforming the baseline (73.25\%) and consistently surpassing other methods.
It also outperforms the more complex E2E augmentation, which involves an iterative process of answering each core sub-question separately before synthesizing them to respond to the original query.
Overall, we find that retrieving chunks that address core sub-questions is highly effective in generating comprehensive answers and can be easily integrated into any existing RAG systems.

\section{Conclusions}
We introduced a novel evaluation framework for Retrieval-Augmented Generation (RAG) systems based on sub-question coverage.
By decomposing complex, open-ended queries into core, background, and follow-up sub-questions, we provided a more refined approach to evaluating RAG-generated responses.
Our results also show that incorporating core sub-questions into RAG systems significantly improves the generation of more comprehensive and accurate answers.

These findings open up new possibilities for evaluating and optimizing RAG systems, particularly for complex, knowledge-intensive tasks.
Additionally, our sub-question-based answer quality metric can serve as a potential reward signal for training end-to-end RAG models, offering a valuable direction for future research and system development.

\section{Limitations}

While our framework for evaluating RAG systems through sub-question coverage offers valuable insights, it has several limitations.
First, the accuracy of automatic sub-question decomposition, although generally reliable, may fail in capturing the full complexity of ambiguous or nuanced questions.
Second, our reliance on GPT-4 for evaluating sub-question coverage may introduce discrepancies compared to human judgment, especially in subjective cases.

Additionally, our approach assumes uniform importance across sub-question types, which may not hold across different domains or contexts.
Furthermore, while we focus on improving core sub-question coverage, this may sometimes overlook the value of background or follow-up sub-questions.
Finally, the computational demands of our method may limit its scalability and real-time application potential.

\bibliography{anthology,custom}

\appendix

\newpage
\onecolumn
\section{GPT-4 Prompts}
\label{sec:appendix-1}
\begin{small}
\begin{longtable}{p{0.9\textwidth}}

\toprule

The prompt used for getting a comprehensive collection of relevant sub-questions for complex questions: \\
\textbf{Decompose the following complex question into a collection of around 20 sub-questions that you think would be relevant to answer the complex question fully.} \\
\\
\textbf{Complex question: \texttt{\$question}} \\
\textbf{Collection of sub-questions:} \\

\midrule[0.03em]

The prompt used for classifying sub-questions into three types: \textit{core}, \textit{background}, and \textit{follow-up} sub-questions: \\
\textbf{Based on the sub-question's relevance and functional role in answering the complex question, classify the sub-question into three types: core, background, and follow-up.} \\
\\
\textbf{The definitions of these three sub-question types are:} \\
\textbf{(1) Core sub-questions:} \\
\textbf{ - They are central to the main topic and directly or partially address the complex question.} \\
\textbf{ - They are crucial for interpreting the logical reasoning of the complex question and provide essential insights required for answering the complex question.} \\
\textbf{ - They often involve multiple steps or perspectives, making them fundamental to generating a comprehensive and well-rounded response to the complex question.} \\
\textbf{(2) Background sub-questions:} \\
\textbf{ - They are optional when answering the complex question, but they can provide additional context or background information that helps clarify the complex question.} \\
\textbf{ - Their primary role is to support the understanding of the main topic by offering supplementary evidence or information, though it is not strictly necessary for addressing the core aspects of the complex question.} \\
\textbf{(3) Follow-up sub-questions:} \\
\textbf{ - They are not needed to answer the complex question.} \\
\textbf{ - They often arise after users receive an initial answer and seek further clarification or details.} \\
\textbf{ - They may explore specific aspects of the response in greater depth, but their answers can sometimes be out-of-scope or beyond the focus of the original complex question.} \\
\\
\textbf{Here are a few examples you can use for reference:} \\
\textbf{\texttt{\$few-shot-examples}} \\
\\
\textbf{Complex question: \texttt{\$question}} \\
\textbf{Sub-question: \texttt{\$sub-question}} \\
\textbf{Type classification:} \\

\bottomrule
\caption{The prompts we use in \autoref{subsec:decomposition}. We omit the output-format controlling prompts for brevity.}
\label{tab:prompt-1}
\end{longtable}
\end{small}
\begin{small}
\begin{longtable}{p{0.9\textwidth}}

\toprule

The prompt used for the automatic measurement of the sub-question coverage: \\
\textbf{You are given a piece of text and a question.} \\
\textbf{Judge if there exists any part of the given text that can answer the question.} \\
\textbf{If you believe the question can be answered, identify the text fragment that answers the question; otherwise, just return ``None''.} \\
\\
\textbf{Here are a few examples you can use for reference:} \\
\textbf{\texttt{\$few-shot-examples}} \\
\\
\textbf{Piece of text: \texttt{\$text}} \\
\textbf{Question: \texttt{\$sub-question}} \\
\textbf{Judgment:} \\

\bottomrule
\caption{The prompt we use in \autoref{subsec:protocol}. We omit the output-format controlling prompt for brevity.}
\label{tab:prompt-2}
\end{longtable}
\end{small}

\newpage
\onecolumn
\section{Samples of Question Decomposition with Three-Way Classification}
\label{sec:appendix-2}
\begin{small}
\begin{longtable}{p{0.9\textwidth}}

\toprule

Sample \#1: \\
Main question: How can human activity affect the carbon cycle? \\
\\
Core sub-questions:
\begin{itemize}
\setlength\itemsep{0.01em}
    \item What human activities contribute to carbon emissions?
    \item How does deforestation affect the carbon cycle?
    \item What role does the burning of fossil fuels play in the carbon cycle?
    \item How do agricultural practices impact the carbon cycle?
    \item What is the effect of urbanization on the carbon cycle?
    \item How do industrial processes alter the carbon cycle?
    \item What is the impact of increased carbon dioxide levels on global warming?
    \item How does the alteration of the carbon cycle affect ocean chemistry?
    \item How can changes in land use affect the carbon cycle?
    \item What are the effects of waste management and landfill operations on the carbon cycle?
    \item How do energy production methods influence the carbon cycle?
    \item How can reforestation and afforestation impact the carbon cycle?
\end{itemize} \\
Background sub-questions:
\begin{itemize}
\setlength\itemsep{0.01em}
    \item What is the carbon cycle and how does it function?
    \item What are the main components of the carbon cycle?
    \item What are the natural sources of carbon emissions?
\end{itemize} \\
Follow-up sub-questions:
\begin{itemize}
\setlength\itemsep{0.01em}
    \item What are the consequences of the carbon cycle disruption on wildlife?
    \item How does the carbon cycle influence climate change?
    \item What are the long-term effects of altered carbon cycles on Earth's ecosystems?
    \item What are some ways to mitigate human impact on the carbon cycle?
    \item What policies can be implemented to reduce carbon emissions?
\end{itemize} \\

\midrule[0.03em]

Sample \#2: \\
Main question: How does reading foster long-term learning? \\
\\
Core sub-questions:
\begin{itemize}
\setlength\itemsep{0.01em}
    \item How does the brain process and store information read from texts?
    \item How does reading comprehension contribute to knowledge retention?
    \item How does the complexity of text affect comprehension and memory retention?
    \item What role does prior knowledge and experience play in reading comprehension?
    \item How does note-taking while reading enhance long-term memory?
    \item What are the neurological benefits of regular reading?
    \item How does reading fiction versus non-fiction impact long-term learning?
    \item How does the frequency of reading affect long-term cognitive abilities?
    \item What role does visualization while reading play in memory retention?
    \item How can reading multiple sources on the same topic enhance understanding and retention?
    \item What are the long-term impacts of reading on academic performance?
    \item How does reading influence critical thinking and analytical skills over time?
    \item What strategies can be employed to improve reading habits for better long-term learning?
\end{itemize} \\
Background sub-questions:
\begin{itemize}
\setlength\itemsep{0.01em}
    \item What is the definition of long-term learning?
    \item What cognitive skills are involved in reading?
    \item How does active reading differ from passive reading?
\end{itemize} \\
Follow-up sub-questions:
\begin{itemize}
\setlength\itemsep{0.01em}
    \item What types of reading materials are most effective for long-term learning?
    \item What are the benefits of discussing or teaching others about what one has read?
    \item What are the effects of digital versus physical reading on learning?
    \item How does age affect the ability to learn from reading?
\end{itemize} \\

\midrule[0.03em]

Sample \#3: \\
Main question: Why is a starving individual more susceptible to infectious disease than a well-nourished individual? \\
\\
Core sub-questions:
\begin{itemize}
\setlength\itemsep{0.01em}
    \item How does malnutrition affect the immune system?
    \item How does protein-energy malnutrition impact immune cell function?
    \item What role do micronutrients play in immune system function?
    \item Which micronutrients are most important for a healthy immune response?
    \item How does deficiency in specific micronutrients affect susceptibility to infections?
    \item How does malnutrition alter the physical barriers of the body that prevent infection?
    \item What is the impact of malnutrition on the gut microbiome?
    \item How does the alteration of the gut microbiome in malnourished individuals affect immune function?
    \item What are the physiological changes in a malnourished body that increase infection risk?
    \item How does malnutrition affect the healing process after an infection?
    \item How does the severity and duration of malnutrition affect the level of increased susceptibility to infectious diseases?
\end{itemize} \\
Background sub-questions:
\begin{itemize}
\setlength\itemsep{0.01em}
    \item What is the definition of malnutrition?
    \item What are the key components of the immune system?
    \item What are the statistics on infection rates in malnourished versus well-nourished populations?
\end{itemize} \\
Follow-up sub-questions:
\begin{itemize}
\setlength\itemsep{0.01em}
    \item What are common infectious diseases that affect malnourished individuals?
    \item How do socioeconomic factors contribute to malnutrition and increased susceptibility to infectious diseases?
    \item What interventions can reduce the impact of malnutrition on susceptibility to infectious diseases?
    \item How effective are nutritional supplements in restoring immune function in malnourished individuals?
    \item What are the long-term effects of childhood malnutrition on adult immune function?
    \item What policies are effective in combating malnutrition and thus reducing susceptibility to infectious diseases?
\end{itemize} \\

\bottomrule
\caption{Samples of question decomposition with three-way classification obtained from \autoref{subsec:decomposition}.}
\label{tab:samples}
\end{longtable}
\end{small}

\end{document}